%% file: egpaper_for_review.tex
\documentclass[10pt,twocolumn,letterpaper]{article}

\usepackage{wacv}
\usepackage{times}
\usepackage{epsfig}
\usepackage{graphicx}
\usepackage{epstopdf}
\usepackage[cmex10]{amsmath}
\usepackage{algorithm,algorithmic}
\usepackage{amssymb, multirow}
\usepackage[caption=false,font=footnotesize]{subfig}
\usepackage{booktabs}
\usepackage{url}
\usepackage{gensymb}

\DeclareGraphicsExtensions{.eps}
\graphicspath{ {images/} }

\usepackage{chngcntr}
\counterwithin{table}{section}
\counterwithin{figure}{section}

\usepackage[pagebackref=true,breaklinks=true,letterpaper=true,colorlinks,bookmarks=false]{hyperref}

\newcommand{\squeezeup}{\vspace{-2.5mm}}

\wacvfinalcopy 

\def\wacvPaperID{55} 

\ifwacvfinal\fi
\begin{document}

\title{A Statistical Approach to Continuous Self-Calibrating Eye Gaze Tracking for Head-Mounted Virtual Reality Systems}

\author{Subarna Tripathi\\
UC San Diego\\
{\tt\small stripathi@ucsd.edu}
\and
Brian Guenter\\
Microsoft Research\\
{\tt\small bguenter@microsoft.com}
}

\maketitle
\ifwacvfinal\thispagestyle{empty}\fi

\input{abstract.tex}
\input{introduction.tex}
\input{relatedwork.tex}
\input{systemoverview.tex}
\input{background.tex}
\input{autocalibration.tex}
\input{results.tex}
\input{conclusion.tex}

{\small
\bibliographystyle{ieee}
\bibliography{egbib}
}

\end{document}

%% file: abstract.tex
\begin{abstract}

We present a novel, automatic eye gaze tracking scheme inspired by smooth pursuit eye motion while playing mobile games or watching virtual reality contents.
Our algorithm continuously calibrates an eye tracking system for a head mounted display. This eliminates the need for an explicit calibration step and automatically compensates for small movements of the headset with respect to the head. The algorithm finds correspondences between corneal motion and screen space motion, and uses these to generate Gaussian Process Regression models. A combination of those models provides a continuous mapping from corneal position to screen space position. Accuracy is nearly as good as achieved with an explicit calibration step. 

\end{abstract}
\squeezeup 

%% file: introduction.tex
\section{Introduction}

Head mounted eye trackers require precise knowledge of the intrinsic parameters of the eye tracking camera, principally focal length and distortion, as well as the extrinsic parameters, the position and orientation of the camera with respect to the display and illumination sources.

Maintaining precise calibration in a consumer HMD device is difficult; laboratory equipment is handled with care but consumer devices may be dropped, sat upon, or left in the sun. Injection molded plastic is not very stiff so even a modest torque or force can distort the headset. Plastics also have a high coefficient of thermal expansion so parts displace relative to each other as the electronics heat up and cool down. 

Most eye tracking systems also require an initial calibration phase, where the user fixates their gaze at target points. This is inconvenient since it must be done every time the headset is put on. The calibration also rapidly drifts due to movement of the headset with respect to the head.

Our algorithm continuously and automatically calibrates the eye tracking system by detecting correspondence between corneal motion and the motion of objects seen in the display. Only very approximate camera intrinsic and extrinsic parameters are needed, so the system is immune to calibration errors caused by small physical distortions of the headset.
Through experimental evaluations, we show that the proposed algorithm can achieve gaze estimation accuracy competitive with that of a calibrated eye-tracker, without any manual calibration.

Our primary contribution is that the proposed approach of real-time eye gaze tracking system is able to auto calibrate through smooth pursuit and allow people to look wherever they want rather than requiring them to track a single object. The system creates and maintains both a global and a local statistical model for eye-gaze tracking and can automatically select the best prediction among these two models. 
The local model can respond much more quickly to changes in headset position and the global model predicts the eye gaze more accurately while the system is in a stable state.

%% file: relatedwork.tex
\section{Related Work} \label{RelatedWork}

Gaze estimation techniques have long been targeted for improving human-computer interaction, and advancement of assistive technologies for impaired. Traditionally, gaze estimation techniques use a system where the user can move in front of a computer screen such as \cite{Morimoto_CVPR_02, IEEE_HMS15}. One of such state-of-the-art remote set-up methods \cite{IEEE_HMS15} exploiting distant camera and nominal head-motion setup reports gaze estimation error about $2.27\degree$.

Alnajar \textit{et al.} \cite{Alnajar_ICCV_13} presented a method to automatically calibrate gaze in an un-calibrated setup by using the gaze patterns of individuals to estimate the gaze points for new viewers without active calibration using saliency maps and reported an average accuracy of $4.3$$^{\circ}$. Other saliency map based auto-calibration methods include \cite{Shi_CVPR_14, Jixu_CVPR_11}.
Hansen \textit{et al.} \cite{Shape_IR_PAMI_10} compared several different approaches for estimating user's point of regard (PoR). 
Moving target based or smooth pursuit based calibration methods such as \cite{pfeuffer13_uist, pursuit_calib_13} exist for remote setup, but remains unexplored for VR mobile gaming.

With wearable devices being more widely used, gaze estimation has also been explored for virtual reality systems and systems with see-through displays. 
In the arena of see-through displays, Pirri \textit{et al.} \cite{Pirri_CVPR_11, Pirri_CVPRW_11} proposed a procedure for calibrating a scene-facing camera's pose with respect to the user’s gaze. In spite of being effective, its dependence on artificial markers in the scene limits the applicability. 

Tsukada \textit{et al.} \cite{Appearance_CodeBook_ICCV_11} presented a single-eye based gaze detection system that leverages an appearance code book for the gaze mapping. The appearance code book is sensitive to the calibration, and assumed to be constant throughout use. Another recent method  \cite{Perra_2015_CVPR} proposed adaptive eye-camera calibration for head-worn devices. The method is able to work well with changes in calibration during the time of use with locally-optimal eye-device transformation by leveraging salient regions. However, their results are  reported on a simulated dataset and human face centered dataset, both are restrictive compared to the real-world cases.

Although state-of-the-art gaze estimation techniques claim to achieve less than one degree error, in practical settings there are several different error sources such as disparity and physiological differences \cite{error_modeling_15} that affect gaze quality over time. 
The challenge of compensating the drift from the initial calibration has been analyzed in \cite{Sugano_UIST15}. Drift analysis is performed on a dataset of natural gaze recorded using synchronized video-based and Electrooculugraphy-based eye trackers of 20 users performing everyday activity in a mobile setting. 

By analyzing the drift, \cite{Sugano_UIST15} proposed a method to automatically self-calibrate head-mounted eye trackers based on a computational model of bottom-up visual saliency such as Boolean Map Saliency, Face Detection, Pedestrian Detection. The high-level saliency detection runs at $17$ frames per second. In spite of being a robust mapping approach based on error analysis on real-world first-person video in practical settings which supports HMD movement, the method is far from being real-time.

Other recent methods which exploit video signals either use appearance-based methods with eye-image as a descriptor to obtain gaze \cite{Tan_WACV_02} position or Shape-based methods by tracking parts of actual eye-anatomy such as corneal reflection, pupil contour, and iris contour \cite{Morimoto_CVPR_02, Scene_MultiSpectral_ECCV12, Gen_theory_Guestrin_2006}. Corneal reflection and pupil contour methods need infrared ray (IR) active illumination \cite{IR_WACV_13, Shape_IR_PAMI_10}. SensoMotoric Instruments (SMI) \cite{SMI-eye-tracker_14} has its prototype eye-tracking VR HMD and showed its eye-tracked Rift at the 2014 Augmented World Expo. That prototype shares the similar kind of architecture as of ours except they use two 3D eye-facing cameras. Their eye-tracking HMD system measures the 3D eye model which is used as input for gaze based interaction schemes. 
The latency is high (between $50$ ms to $100$ ms) for applications such as foveated rendering. 

Apart from being computationally-heavy geometric model estimator, the system is not truly calibration-free. The utility contains a default explicit one-point calibration and a three-points calibration procedure which asks the user to look at each dot and pressing a button. In spite of being highly accurate ($0.5$ to $1\deg$) at the cost of latency and power, the tracking accuracy is quite sensitive to HMD placement. Even a little pressure on the HMD without really moving it, can cause noticeable shift and the system needs to go through the explicit calibration utility again as per users feedback \cite{SMI-eye-tracker_blog}.

Very recently convolutional neural network based methods such as \cite{nips15_recasens, cvpr2016_Khosla} are proposed. The authors collect a new dataset containing a set of images as well as annotation of the gaze of each person inside an image. In spite of providing an interesting research frontier, these methods are not yet foveated rendering ready since the angular errors are larger than $10\degree$. 

Our proposed method of real-time eye gaze tracking system is able to auto calibrate and allow people to look wherever they want rather than requiring them to track a single object. The proposed approach relies on simple and efficient method of exploiting smooth pursuit for appropriate visual signals present in mobile games or Virtual Reality video content. 

%% file: systemoverview.tex
\section{System Overview} \label{systemOverview}

\begin{figure}[!t]
\centering
\includegraphics[width=2.5in]{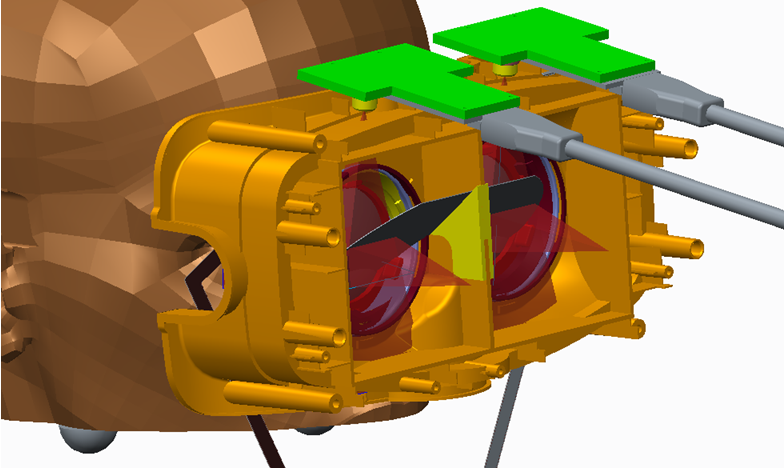}
\caption{Cutaway CAD view of the prototype HMD made from a modified Oculus Rift Dk2. We had the Oculus shell professionally 3D scanned and changed the CAD file to insert custom prototype high speed cameras and an IR reflecting hot mirror. Each camera (the green T shapes) looks down from the top of the HMD and sees the reflected image of the eye in the hot mirror. The user looks through the hot mirror to the display, not shown in this figure. The user cannot see through the front of the HMD. (best viewed in color)}
\label{HMDCAD}
\end{figure}

The eye tracking prototype is a modified Oculus Rift Dk2, as shown in fig. \ref{HMDCAD}. Two custom prototype high speed cameras are mounted on the top of the headset and see the eye reflected in a hot mirror tilted at 67 degrees. The eye is seen through the lens of the Oculus. 

The prototype cameras use a custom image sensor developed in house. It has 13um square responsive pixels with a unique ADC per pixel architecture. The imager is capable of high-frame rates, up to 1000 fps with an exposure time of 500us, and low-power operation ranging from 60mW at 1000 fps to 3mW at 60fps.

All except one of the prototype cameras failed before we could gather results so the 
system used for this paper has a single camera. 
The left eye looks at the display through the Oculus lens and the right eye is tracked by the camera. A black screen is presented to the right eye to avoid vergence problems.

\begin{figure}[!t]
\centering
\includegraphics[width=2.5in]{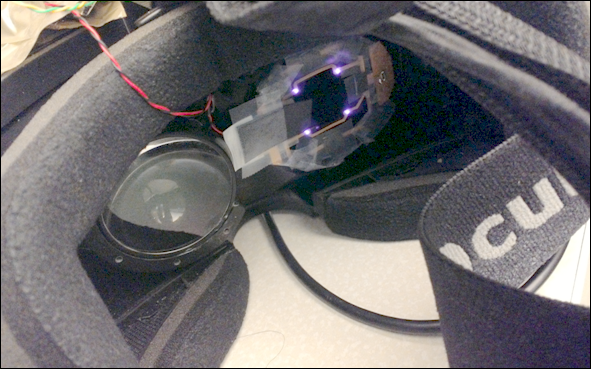}
\caption{Closeup of the interior of the prototype showing the led illuminator array.}
\label{IL}
\end{figure}

An infrared illuminator shines on the eye, creating bright glints on the surface of the cornea. The closeup image, fig. \ref{IL}, shows the IR led illuminator array more clearly. These glints are used to compute the center of the three-dimensional corneal sphere based on similar principles described in  \cite{Cornea_center07}.
The x,y,z coordinates of the corneal center are the input to the learning algorithm. 

\begin{figure}[!t]
\centering
\includegraphics[width=1.6in]{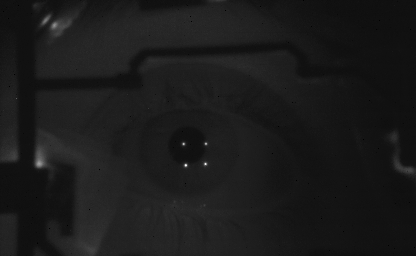}
\includegraphics[width=1.6in]{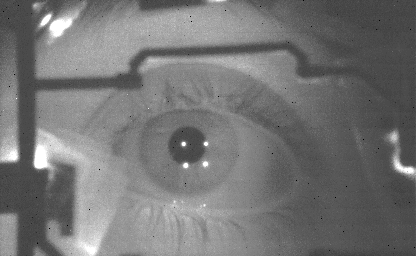}
\caption{The eye as seen by the eye tracking camera. On the left is the raw image. Because we track only glints the exposure is very dark. This reduces illuminator power and allows higher frame rates at any given power. On the right the brightness level of the image has been raised to show more detail.}
\label{capture}
\end{figure}

The camera view of the eye is shown in fig. \ref{capture}. Because only glints are tracked the exposure is dark, as seen in the raw camera image on the left. On the right the brightness levels have been raised so the eye and illuminator circuitry are both visible.

Tracking glints alone, rather than also tracking the pupil, is attractive for several reasons. Far less illumination power is required, a big plus for mobile HMD's. The brightness of the glints also makes it possible to capture at very high frame rates. Our system works at up to 500Hz. Computation is also low; bright spots are much easier to track than the pupil and many fewer pixels have to be processed. There is far less variation across people in corneal reflectance than in the contrast between pupil and iris so tracking should be more robust.

%% file: background.tex
\section{Background} \label{background}

The regression function which maps the cornea location in the camera coordinate system to screen space gaze location can either be model-based or statistical in nature. In case of a parametric model, a common practice is to fit a bi-quadratic function to estimate the mapping where the corneal location and corresponding screen gaze points are collected by asking the user to fixate the gaze at some predefined screen points while keeping the head as still as possible. 

We explore on Gaussian Process Regression (GPR) for a statistical model. Here each observation $\mathbf{y}$ can be thought of as related to an underlying function $f(\mathbf{x})$ through a Gaussian noise model as:
\begin{equation}
\label{GN_model}
\mathbf{y} = f(\mathbf{x}) + \mathcal{N}(0,{{\sigma}_{n}}^2)
\end{equation}
One observation is related to another through the \emph{covariance function}, $\mathcal{K}(\mathbf{x,x'})$. We fold the noise into $\mathcal{K}(x,x')$ as in \cite{GPR_08}:
\begin{equation} \label{GPR_model}
\mathcal{K}(\mathbf{x,x'}) = {{\sigma}_f}^2\exp[\frac{\mathbf{-(x-x')}^2}{2l^2}] + {{\sigma}_{n}}^2\delta(\mathbf{x,x'})
\end{equation}
where $\delta(\mathbf{x,x'})$ is the Kronecker delta function. Thus, given $n$ observations $\mathbf{\bar{y}}$, our objective becomes predicting  $\mathbf{y_{*}}$, and not the actual value of $f_{*}$. 
To prepare for the GPR, we calculate the covariance function between all possible combinations of the points. Extreme off-diagonal elements tend to be zero when $\mathbf{x}$ spans a large enough domain. For, 9-point baseline calibration setup, the covariance function takes the following form:

\[ \label{baseline_K}
\mathcal{K}=
  \begin{bmatrix}
   k(\mathbf{x_1,x_1}) & k(\mathbf{x_1,x_2}) & \cdots & k(\mathbf{x_1,x_9}) \\
   k(\mathbf{x_2,x_1}) & k(\mathbf{x_2,x_2}) & \cdots & k(\mathbf{x_2,x_9}) \\
   \vdots  & \vdots & \ddots & \vdots \\  
   k(\mathbf{x_9,x_1}) & k(\mathbf{x_9,x_2}) & \cdots & k(\mathbf{x_9,x_9})
  \end{bmatrix}
\]
According to the key assumption in GP modeling that the data can be represented as a sample from a multivariate Gaussian distribution, we have

\begin{equation}
\label{multivariate_gaussian}
\begin{pmatrix}\mathbf{\bar{y}} \\ \mathbf{y_{*}}\end{pmatrix} \sim \mathcal{N}( 
\mathbf{0}, 
\begin{pmatrix} \mathcal{K} & {\mathcal{K}_{*}}^T \\  \mathcal{K}_{*} & \mathcal{K}_{**} 
\end{pmatrix}) 
\end{equation}

The conditional probability $\mathbf{y_{*}}|\mathbf{\bar{y}}$ follows a Gaussian distribution:

\begin{equation}
\label{conditional_probability}
\mathbf{y_{*}}|\mathbf{\bar{y}} \sim \mathcal{N}(\mathcal{K}_{*}K^{-1}\mathbf{\bar{y}}, \mathcal{K}_{**} - \mathcal{K}_{*}K^{-1}{\mathcal{K}_{*}}^T)
\end{equation}

Thus, the best estimate for $\mathbf{y_{*}}$ is the mean of the distribution  $\mathcal{K}_{*}K^{-1}\mathbf{\bar{y}}$ and the uncertainty in the estimate is captured in its variance: $var(\mathbf{y_{*}}) =  \mathcal{K}_{**} - \mathcal{K}_{*}K^{-1}{\mathcal{K}_{*}}^T $ 

where, $\mathbf{x_{*}}$ is the input test point, $\mathcal{K}_{*} = [k(\mathbf{x_{*}, x_{1}}), k(\mathbf{x_{*}, x_{1}}), ..., k(\mathbf{x_{*}, x_{9}})]$ and $\mathcal{K}_{**} = k(\mathbf{x_{*}, x_{*}})$. \\
We note that the maximum allowable covariance defined as ${\sigma_{f}}^2$ should be high for functions which cover a broad range on the $\mathbf{y}$ axes. Thus, the values of the hyper-parameters i.e. ${\sigma_{f}}^2$, ${\sigma_{n}}^2$ and the kernel width, $l$ should be selected based on the nature of the input and output data. For the baseline system, we use mean-centered unit-norm data distribution and select the values of these hyper-parameters accordingly.

%% file: autocalibration.tex
\section{Auto-Calibration} \label{autoCalibration}

The proposed auto-calibration is about learning the mapping from corneal location $\mathbf{x}$ ($x_{c},y_{c},z_{c}$) to screen gaze location $\mathbf{y}$ ($x_{g},y_{g}$) through Gaussian Process Regression. However, unlike explicit 9-point calibration method the correspondence between corneal location and screen gaze point data is not available directly. In addition, the system needs to be able to recover from minor HMD movement with respect to the head. 

A typical scenario of mobile video games where 2D icons are moving over the screens with perceivable speed, smooth  pursuit becomes the most significant eye motion while playing those games. Our auto-calibration algorithm exploits smooth pursuit eye movement to dispense with the need of explicit calibration and is able to continuously calibrate the wearable system. 

\subsection{Tracklet}
We have the corneal location measurement as explained in section  \ref{systemOverview}. To identify the corresponding object on screen, we find trajectory similarity between cornea and one of the multiple moving objects on the screen space. Since it is natural for a user to follow a moving object for some tangible duration of time and then change gaze and follow another one, we look for similarity between corneal trajectory and screen-space motion trajectories within a temporal neighborhood. We define temporal locality in terms of tracklets. A query tracklet is a collection of consecutive \emph{n} samples in corneal trajectory. 
In our experiments, $n=40$ to $50$. Apart from the trajectory appearance similarity between the query and candidate tracklets, an informed guess about the possible spatial locations of screen-space candidates leads to computational tractability and improved eye tracking accuracy. We describe the tracklet matching strategy in sub-section \ref{trackletMatching}.   

\subsection{Tracklet Matching} \label{trackletMatching}
\begin{figure*}
\begin{center}
	\includegraphics[width=3.0in]{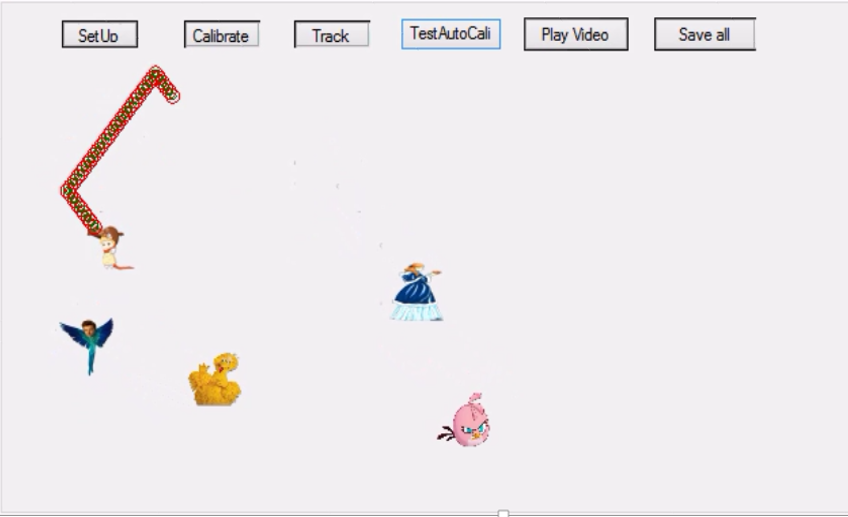}    
	\includegraphics[width=3.0in]{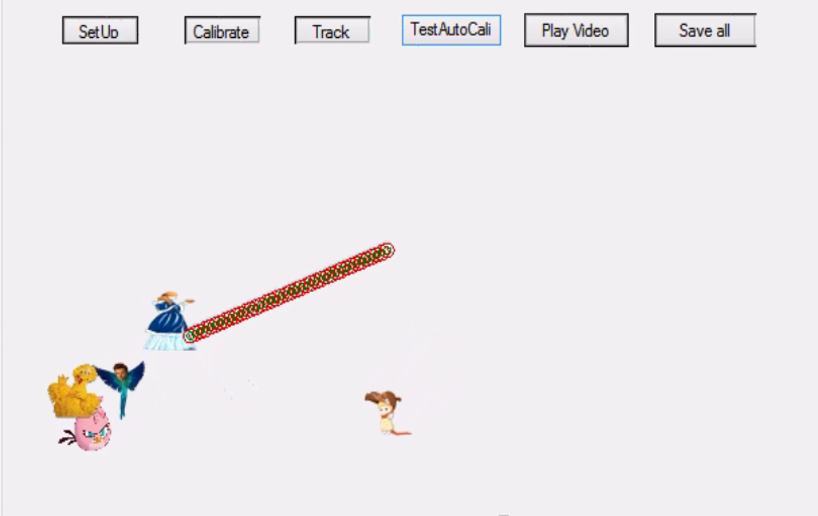}
\end{center}
   \caption{Visualization of best candidate tracklet selection. At two different time instant, the best matched corresponding screen-space tracklets are shown. The left one corresponds to following the path of flying baby and the right one corresponds to following the path of princess for about half a second in each case. Green circles are the selected screen-space tracklet and the red circles demonstrate the correctness of the local model learned on the those data-pairs i.e.\ predicting gaze point for those associated corneal locations. (best viewed in color)}
\label{fig:tracklet_match_viz}
\end{figure*}

In the process of learning the mapping between $3D$ mapping from corneal location to $2D$ screen-space location, 
we observe that the 2D corneal trajectory (assuming constant depth of corneal sphere with respect to the camera) and the motion of the object the user follows on screen space has appearance similarity in a small spatio-temporal neighborhood. In the beginning, without any knowledge of the possible mapping from corneal location to screen gaze points, we rely on discrete normalized velocity similarity between the corneal trajectory and the best matched object's motion. 
Limited number of independently moving 2D icons over screen space makes it possible to evaluate candidate tracklets efficiently. Figure \ref{fig:tracklet_match_viz} show some matched tracklets. 

The horizontal and the vertical direction of motion is assumed to be independent. 
Tracklet matching algorithm takes the inputs of synchronized time series of horizontal and vertical corneal locations and the coordinates of all objects on the screen space and outputs the single object whose trajectory is most similar to that of the eye or nothing if no object is being followed. The process is continuously performed over small temporal windows. With informed guess on search-space, directional independence hypothesis, and smooth pursuit in small temporal window - initialization of tracklet-matching reduces from complex 2D/3D trajectory shape matching to finding minimum euclidean distances between horizontal and vertical velocity vectors independently. \\

\textbf{Initialization} Let, $\mathbf{C_{X}}$ and $\mathbf{C_{Y}}$ denote the corneal X and Y position vectors for the query tracklet in temporal order. We begin with normalizing $\mathbf{C_{X}}$ and $\mathbf{C_{Y}}$ so that the sum of the individual elements in both the vectors becomes unity. All the candidate screen space tracklets, $\mathbf{S^i_{X}}$ and $\mathbf{S^i_{Y}}$ are similarly normalized where $i$ denotes the index of the candidate tracklet. 
The direction of the horizontal component of the corneal velocity vector with respect to the eye-facing camera coordinate system is opposite to that of the corresponding object's horizontal velocity. Therefore, direction of corneal horizontal velocity is reversed before tracklet appearance matching process. Let, $D^j_{X}$ be the euclidean distance between normalized sign-reversed vector of horizontal motion and $j$-th screen space trajectory's horizontal component and $D^j_{Y}$ be the euclidean distance between normalized corneal vertical motion and $j$-th screen space trajectory's vertical component for valid temporal indices. Then the appearance similarity $S_{j}$ with $j$-th candidate tracklet is calculated as: 

\begin{align} 
\label{eqn_similarity}
\begin{split}
{D^j_{X}} = \Sigma(\mathbf{-NC_{X}} - \mathbf{NS^j_{X}})
\\
{D^j_{Y}} = \Sigma(\mathbf{NC_{Y}} - \mathbf{NS^j_{Y}})
\\
D_{j} = \sqrt{{{D^j_{X}}}^2 + {{D^j_{Y}}}^2}
\\
S_{j} = \exp({- D_{j}/{s_{n}}})
\end{split}
\end{align}

where prefix $N$ denotes normalized vectors and $s_{n}$ is the cardinality of the query tracklet. 
If the horizontal coordinate value of the object gets bigger when moving to the left or right of the screen, horizontal location of the eye should also increase. This implies that the screen should be located orthogonal to the line of sight which is mostly true for HMD setup. However, some relative movement of HMD can invalidate the constraint and there arises the need of tracklet matching with other efficient method.\\

\textbf{After Initialization}: The candidate with the highest similarity is the one we start learning the GPR (\ref{conditional_probability}) where the input is 3D-corneal trajectory and the output is 2D screen-space gaze locations. 
After the first set of samples collected through initial temporal window, we obtain the first set of corresponding corneal ($\bar{\mathbf{x}}_{1}$) and screen gaze data points ($\bar{\mathbf{y}}_{1}$). 
Thus, we have initial estimate of the covariance matrix ($\hat{\mathcal{K}}_{1}$) of the first local GPR model. The initial appearance similarity based matching requires alignment with line-of-sight \textit{i.e.} at every new frame we update the local mapping model with the newly found valid corneal and screen-space data-pair using the \emph{so far} learned GPR model. Updating the model means updating $\bar{\mathbf{x}}_{i}, \bar{\mathbf{y}}_{i}$, and $\hat{\mathcal{K}}_{i}$; where $i$ denotes the index of the local GPR model. The term local model signifies the locality of space and time which a small tracklet satisfies. 

If the distance between the gaze prediction on screen space ($\mathcal{K}_{i*}K^{-1}_{i}\mathbf{\bar{y}}_{i}$) using the continuously learned GPR model and the best screen space candidate is small enough, we use that data-pair to update the corresponding local GPR model. 
The local model is able to quickly respond to the changes of the relative pose between the head and the headset.

\subsection{Local GPR Models}
The pursuit is initiated with every moving signal and under most natural circumstances, a user looks at or follows different objects in different time instant. As every tracklet is localized in space-time, it makes sense to maintain multiple local GPR models to ensure better space coverage for minimizing prediction uncertainty. As GPR involves matrix inversion, using all corresponding data-pairs for all the matched tracklets is prohibitive from computation and memory point of view.   
We perform the continuous local mapping model update and maintain them in a circular buffer fashion. Every buffer can be thought of as pointing to a block of varying size which is less or equal to the tracklet sample size in the overall system covariance matrix. In the absence of prior information on all the data at once, we approximate the system covariance matrix as a block diagonal matrix assuming each tracklet is independent. The maximum size of the system matrix $\mathcal{K}$ could at most be the number of points present in a tracklet ($n$) times the number of local GPR models ($m$) maintained. We use the value of $m$ between $8$ to $15$ for our experiments and observed that they yield similar accuracy. Size of each local block can vary based on the prediction accuracy based validity criteria. The circular buffer implementation replaces the oldest block in the system covariance matrix with the newest one.     

\begin{equation} \label{eqn:system_block_diagonal}
\mathcal{K} =
\begin{pmatrix}
\begin{bmatrix}&\hat{K}_{1}& \end{bmatrix} \\
\multicolumn{2}{c}{$\upbracefill$}&\\
\multicolumn{2}{c}{\scriptstyle n_1 \times n_1}&\\
         & &  \begin{bmatrix}&\hat{K}_{2}&\end{bmatrix}   \\
         & &  \multicolumn{2}{c}{$\upbracefill$}&\\
		 & &  \multicolumn{2}{c}{\scriptstyle n_2 \times n_2}&\\
         &  & & \ddots     \\
         & & & & \begin{bmatrix}& \hat{K}_{m} \end{bmatrix}  \\
         & & & & \multicolumn{3}{c}{$\upbracefill$}&\\
		 & & & & \multicolumn{3}{c}{\scriptstyle n_m \times n_m}&\\      
         \\
 \end{pmatrix} 
 \end{equation}
 
Our intuition is that the best so far model learned in the far past can not compensate for a recent possible change in relative head to HMD pose. 
System's covariance matrix including all local models looks like equation \ref{eqn:system_block_diagonal}. It always maintains the most recent $m$ number of local models each of which could be of different sizes during online learning. Their respective covariance matrices are denoted by $\hat{K}_i$, where $i\in [1,m]$. 

\subsection{Global GPR Model}
Tracklet independence assumption yields a block-diagonal system covariance matrix which is computationally tractable but not necessarily a good approximation for actual system covariance where data correspondences are spread across a wide range of screen space. Hence, the local GPR based prediction might not be highly confident in the entire screen space region. Thus, we want to have a set of sparse samples of cornea-screen data pairs over the screen space which are valid for relatively longer duration of time. The GPR model learned using those samples is called global model. 

For encouraging space-diversity, we divide the screen-space into non-overlapping rectangular grids and use only limited number of best data pairs seen from all matched tracklets per grid. In our experiments, we use $6\times4$ grids 
and maintain at most $4$ best data pairs per grid. Thus, at any moment the global GPR model deals with inverting a matrix of at most $96\times96$ size. As the covariance matrix is symmetric and we get data in incremental fashion, we can employ incremental matrix inversion strategy for updating both the global model and the local models. The covariance matrix for the global model, $\hat{K}_g$, is not block-diagonal and captures the essence of eye-screen transformation capable of performing a better prediction through interpolation. 



\subsection{Overall Framework}
We initialize the first local GPR model by establishing the correspondence between corneal and screen-space tracklet by trajectory appearance similarity. Then, we keep updating the first local GPR model as we keep getting more on-the-fly corneal and screen-space correspondences. The covariance matrix corresponding to the first matched-tracklet becomes the first block of the block-diagonal system covariance matrix. We then continue matching tracklets based on the model learned so far.   
We keep $m$ of such local GPR models created by good correspondences from the last $m$ matched tracklets. 
At every frame, the current local GPR model is updated with the new corresponding corneal and screen-space tracklet data pair. The space-diverse sampling of all data-pairs is continuously performed for creating a global model.

Online gaze detection involves two gaze predictions: one from the best local model and another from the global model. Once the samples for learning the global model tend to have more space-diversity than the current local model, the prediction from the global model obtains higher confidence in any region of the screen than the one from the local models. Local models are better responsive to relative eye-HMD movement. 
The prediction from global model is more confident for stable-HMD, specially during saccadic corneal motion. When prediction from global model suddenly becomes significantly less confident compared with the local model, we consider it as a notable relative movement between the head and the HMD. Therefore, we start the process of recreating global GPR model. The test time gaze prediction becomes the selection between the local model-based prediction and the global model-based prediction.


\begin{figure*}[!t]
\centering
\subfloat[Local model prediction better than the one from global model ]{\includegraphics[width=3.0in]{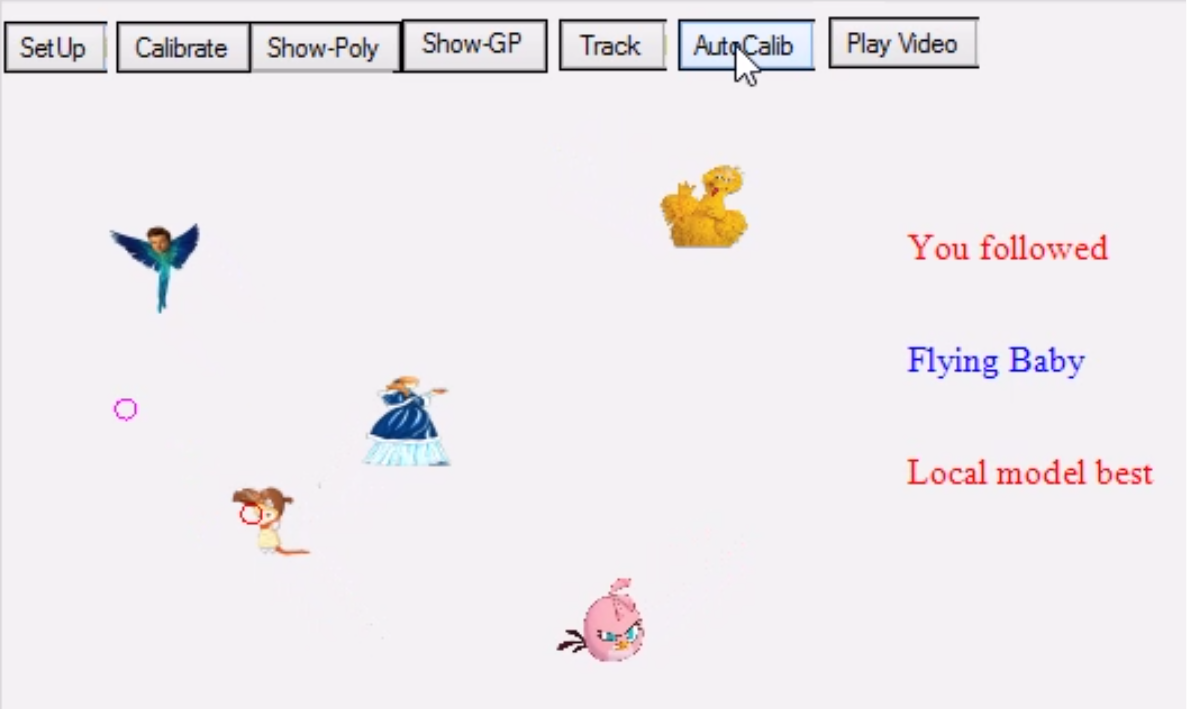}%
\label{fig_first_case}}
\hfil
\subfloat[Global model prediction better than the one from local model]{\includegraphics[width=3.0in]{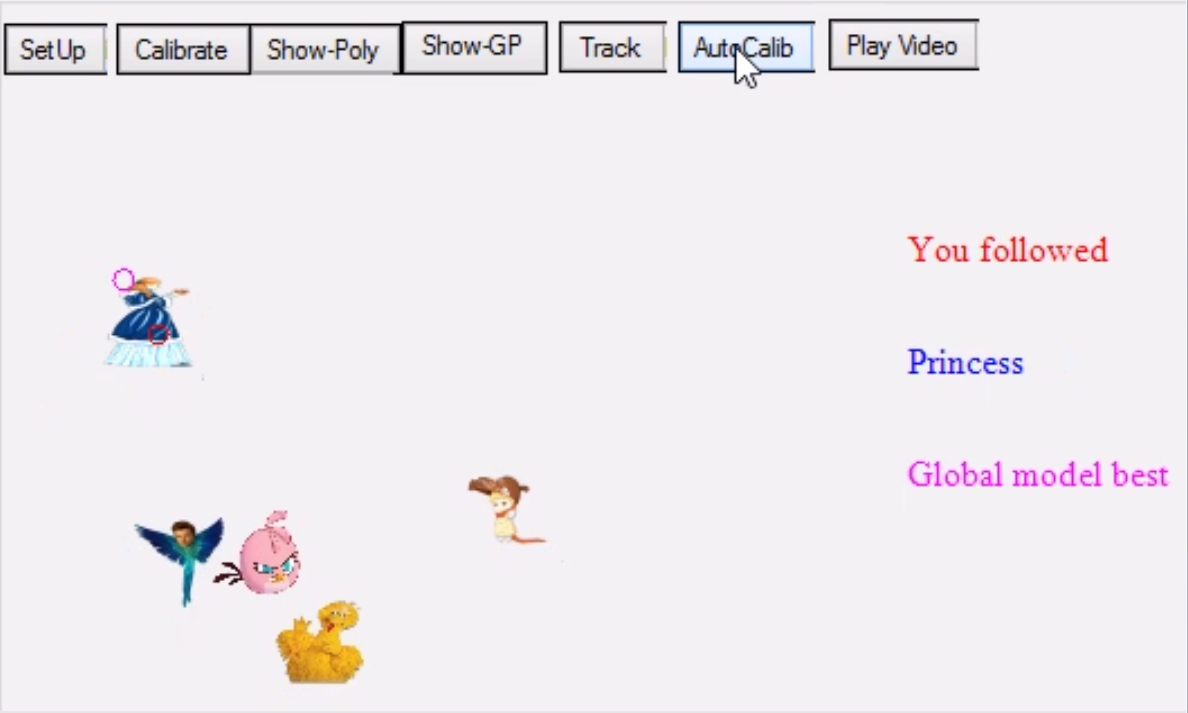}%
\label{fig_second_case}}
\caption{These screen-shots show that there are two test time gaze predictions generated - one from the global model which is shown in Magenta and another one from the best of all local models which is shown in Red. The final estimated gaze point is selected based on the uncertainty of predictions from both the global and local model. (best viewed in color)}
\label{fig:local_vs_global_selection}
\end{figure*}
The gaze estimate $y_{l*}$ from the local model and its uncertainty $var(y_{l*})$ are calculated as:
\begin{align} 
\label{eqn: best_local_pred}
\begin{split}
j = \arg\min_{i \in C}(\hat{K}_{i**} - \hat{K}_{i*}\hat{K}_i^{-1}{\hat{K}_{i*}}^T)
\\
y_{l*} = \hat{K}_{j}{*}\hat{K}_{j}^{-1}\mathbf{y}
\\
var(y_{l*}) = \hat{K}_{j**} - \hat{K}_{j*}\hat{K}_j^{-1}{\hat{K}_{j*}}^T
\end{split}
\end{align}
where $C$ is the set of indices of all local models considered. 

And, the gaze estimate, $y_{g*}$, from the global model and its uncertainty, $var(y_{g*})$, are calculate as :
\begin{align} \label{eqn: global_pred}
\begin{split}
y_{g*} = \hat{K}_{g}{*}\hat{K}_{g}^{-1}\mathbf{y}
\\
var(y_{g*}) = \hat{K}_{g**} - \hat{K}_{g*}\hat{K}_g^{-1}{\hat{K}_{g*}}^T
\end{split}
\end{align}

If the prediction from the global model has more uncertainty $var(y_{g*})$ than the uncertainty $var(y_{l*})$ of the best local model, either due to less number of training samples generated so far or due to relative head-HMD movement, the global model is disposed and created again.

Final gaze prediction, $y_*$ becomes:
\begin{equation} \label{eqn: gaze_pred}
y_{*} = \begin{cases}
y_{l*} & \text{if $var(y_{g*}) > var(y_{l*})$;}\\
y_{g*} & \text{otherwise.}
\end{cases}
\end{equation}

On the fly selection between prediction from local and global model has been visually validated in the screen-shots as shown in figure \ref{fig:local_vs_global_selection}. 

While creating the covariance matrix i.e.\ the individual blocks in equation \ref{eqn:system_block_diagonal}, we do not consider mean-centered unit-norm distribution since estimating screen-space mean value is difficult for localized data points of a single tracklet. On the contrary, the global model can use mean-centered unit-norm data distribution by virtue of space-diverse sampling of data points. The different nature of the data distribution imposes the need of using different values for the hyper-parameters. Thus, the estimated standard deviation from local and global model has difference in scale which is compensated before comparing the uncertainty of gaze prediction from local and global model (eq. \ref{eqn: gaze_pred} ).  


%% file: results.tex
\section{Experimental Results} \label{results}
We evaluate both the efficiency of the explicit calibration-based GPR baseline and the proposed continuous auto-calibration method. The explicit calibration requires a user to look at a set of predefined targets as shown in Fig \ref{fig:Baseline_eval} to establish a mapping from cornea location to gaze positions in the user’s visual scene. 
In laboratory settings, this explicit method serves as the golden standard for eye gaze tracking evaluation.

\subsection{Evaluation Method : 9-Point Baseline}

\begin{figure}[!t]
\centering
\includegraphics[width=3.5in]{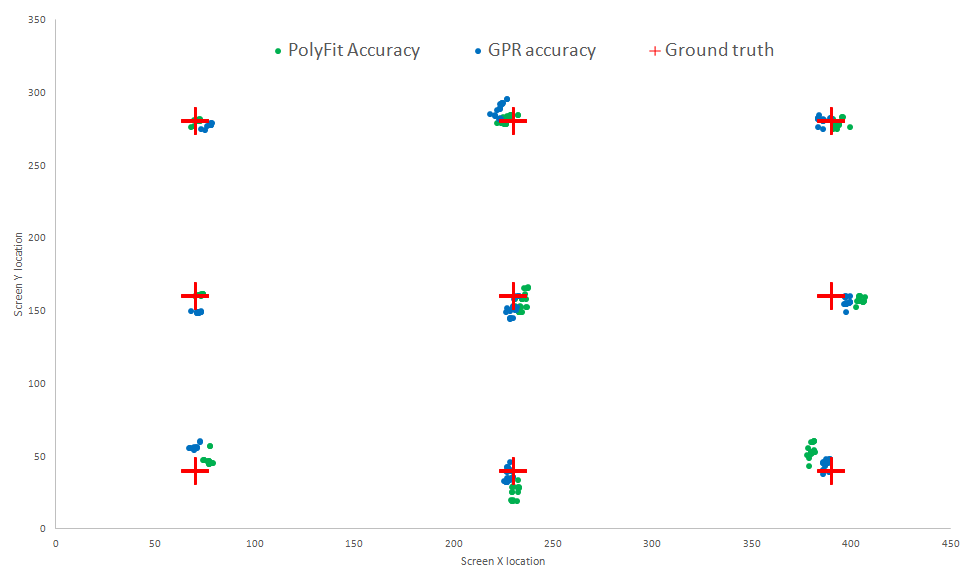}
\caption{9-Point baseline calibration evaluation. Both the polynomial model and the GPR based methods exhibit similar accuracy. (best viewed in color)}
\label{fig:Baseline_eval}
\end{figure}

We learn the best baseline model assuming the transformation function is going to be the same for the entire duration of use. 
In the current HMD setup, the values of corneal locations in the camera coordinate system vary from $-2.5$ to $+2.5$ millimeters in horizontal ($x$) direction, $-2.5$ to $+3.5$ in the vertical ($y$) direction and $100$ millimeter or more in $z$-direction for different users and different relative poses. In case of baseline GPR, we use corneal ($x_c,y_c,z_c$) location as input and corresponding screen-gaze ($x_s,y_s$) as labels for learning the regression. The range of the values for screen space gaze locations goes from $0$ to $500$ in $x$ direction and $0$ to $300$ in $y$-direction in the current hardware. We use mean-centered unit-variance data for learning and evaluating the model. The kernel width we use is $10$. The hyper-parameters in the GPR model such as the maximum allowable covariance and noise parameters are $1.2$ and $0.01$ respectively.

Additionally, we perform bi-quadratic polynomial fitting. 
We see that both the bi-quadratic mapping and Gaussian Process Regression based transformation learning exhibit similar gaze prediction accuracy for the current system. Figure \ref{fig:Baseline_eval} shows the evaluation for baseline methods using bi-quadratic and GPR visually. GPR additionally provides confidence of each gaze point prediction.

\subsection{Evaluation of Continuous Auto-calibration} \label{evaluations}

\begin{figure}[!t]
\centering
\includegraphics[width=3.5in]{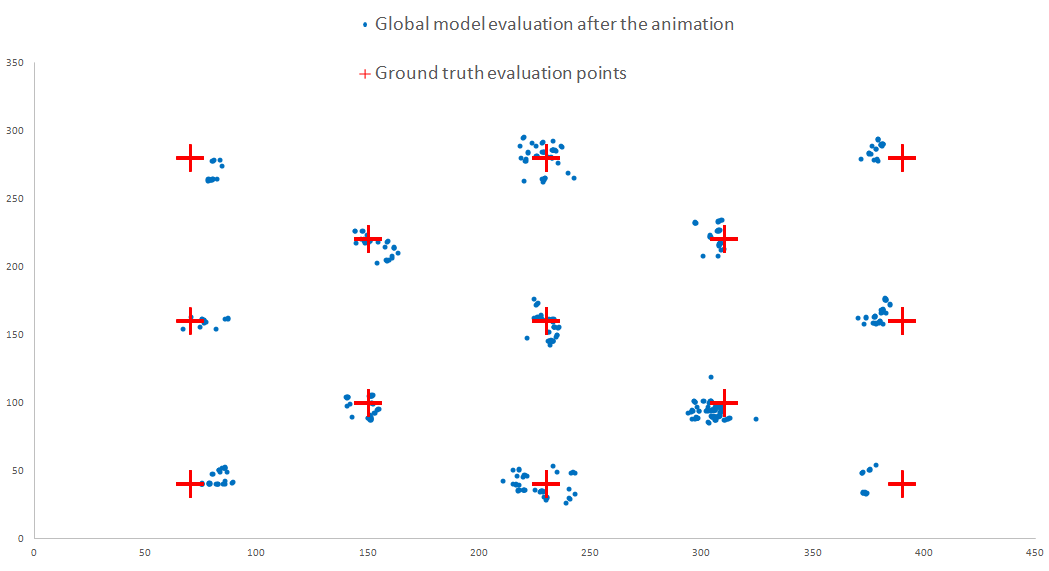}
\caption{13-point evaluation for the continuous auto-calibration method after the game. (best viewed in color) }
\label{eval_autocalib}
\end{figure}

Fig \ref{eval_autocalib} shows the visual $13$-point evaluation result of the proposed auto-calibration. Table \ref{table:gaze_pred_eval_VR} shows the average gaze prediction error for VR systems. We are aware of only one commercial system SMI \cite{SMI-eye-tracker_14} which is made for and evaluated on virtual reality setup. In spite of being highly accurate, \cite{SMI-eye-tracker_14} has high latency and the tracking is significantly sensitive to HMD placement. 
SMI is not fully calibrating free. The rest of the entries correspond to the baseline methods in the same hardware with explicit calibration procedure. Given the fact that there are error sources such as chromatic aberration, absence of two cameras and temporal synchronization between the camera and the display in the existing hardware, the results produced with this hardware are promising. The proposed continuous auto-calibration method performs almost as good as the baseline active calibration procedure. 

\begin{table}[h] 
\centering
\begin{tabular}{lll}
\multicolumn{3}{c}{\textbf{Comparison with VR systems and Baselines}} \\ 
\midrule
    Methods & Setup & Error (\degree) \\ \midrule
    SMI \cite{SMI-eye-tracker_14} & VR & 0.5 to 1\\ 
    9-pt calib, chin rest (Poly fit) & VR & 0.75 \\
    9-t calib, chin rest (GPR) &VR & 0.776\\
    9-pt calib, no chin rest (Poly fit) & VR &  1.223\\
    9-pt calib, no chin rest (GPR) & VR & 1.248\\
    \midrule
    Proposed. Auto-calib, 13-pt eval & VR & $\mathbf{1.822}$\\   
\bottomrule
\end{tabular}
\caption{Comparison with VR systems.}
\label{table:gaze_pred_eval_VR} 
\end{table}

We also compare the results of our method with other auto-calibration methods reported for wearable devices (Table \ref{table:gaze_pred_eval_auto_calib_AR}). Self-calibrating eye gaze tracking methods \cite{Perra_2015_CVPR, Sugano_UIST15} for head-worn devices deal with see-though displays with scene-facing cameras which exploit 2D/3D saliency maps. Though not specifically meant for VR, methods involving the see-through display system still provide a good reference for our approach.  

The accuracy of our method applied to mobile games is comparable to the performance of \cite{Perra_2015_CVPR} on artificial dataset. Additionally, in comparison to 4-5 fps speed as reported in  \cite{Perra_2015_CVPR}, our method is able to run in real-time. 

Almost immune to HMD movement and amenable to practical settings, the fully-automatic system in \cite{Sugano_UIST15} is far from being real-time due to the associated high-level saliency detection which reportedly runs at most 17 frames per second. Since our approach tries to find correspondences between corneal and screen-space object tracklets, it allows the selection of very small targets which could be difficult for other saliency based methods. 

%
%


\begin{table}[h] 
\centering
\begin{tabular}{lll}
\multicolumn{3}{c}{\textbf{Comparison with auto-calibration AR systems}} \\ 
\midrule
    Methods & Setup & Error (\degree) \\ \midrule
    \cite{Perra_2015_CVPR} on human dataset & AR & 0.168 \\ 
    \cite{Perra_2015_CVPR} on simulated dataset & AR & 1.494 \\
    \cite{Sugano_UIST15} : immune to HMD movement & AR &  $5$ to $10$\\
    \midrule
    Proposed cont. Auto-calib & VR & $\mathbf{1.822}$\\    
\bottomrule
\end{tabular}
\caption{Comparison with auto-calibration AR systems.}
\label{table:gaze_pred_eval_auto_calib_AR} 
\end{table}

There are other self-calibrating eye-tracking systems available in the literature (Section \ref{RelatedWork}) that are applied in remote eye distance scenario. The detailed performance analysis of such methods are outside the scope of this work which deals with wearables. However, from the perspective of technical relevance \textit{i.e.} smooth pursuit based calibration, we summarize some methods. The typical pursuit calibration method such as \cite{pfeuffer13_uist} deals with sampling in a specific and restrictive way, by displaying a single smoothly moving target to the user. \cite{pursuit_calib_13} can deal with only linear or circular object trajectories. Our unified statistical model-based matching and prediction dispenses with the linear trajectory assumption. GPR by its inherent nature is able to match and predict any kind of trajectory shape given the eye-object data-pairs which is incrementally estimated in our tracklet matching approach. 

%% file: conclusion.tex
\section{Conclusion} \label{Conclusion}
We propose an effective continuous auto-calibration for eye tracking in head mounted displays using smooth pursuit
eye movement.
In most natural circumstances, smooth pursuit is generated from signals commonly present in mobile games in virtual reality environment.
For automatically learning and updating the eye to screen mapping in real-time, Gaussian Process Regression models are learned through corneal and screen-space trajectory matching. 
A set of continuously updated local GPR models which are valid mappings on small temporal windows respond to HMD movement faster than the global GPR model which tracks over the entire screen space with higher confidence. 
The statistical model based unified matching and prediction dispenses with the need of linear trajectory assumption. 
A combination of local and global model enables real-time consistent automatic eye tracking system. 
Our results show that the proposed system achieves nearly as good performance as explicit calibration but without imposing any constraints such as fixating gaze as per given instruction.
The method is also immune to minor relative head-HMD movement.

The method can also be extended for the VR world which can enhance user experience. For VR videos, we have forward and/or backward optical flow data at pixel level. The generation of candidate tracklets can potentially be performed through linking the optical flow as in \cite{Rubinstein_BMVC_12} to create pixel level trajectories. 
A space-diverse sampling of all the tracklets with a bias towards horizontal motion can limit the search space over all the candidate tracklets with some initial guess
Our future work lies in exploring continuous auto-calibration for such realistic VR video data.